\documentclass{article}

\def\fig {Figure~}

\usepackage{authblk}

\usepackage[verbose=true,letterpaper]{geometry}
\AtBeginDocument{
  \newgeometry{
    textheight=9in,
    textwidth=6.5in,
    top=1in,
    headheight=14pt,
    headsep=25pt,
    footskip=30pt
  }
}

\usepackage{times}
\usepackage{helvet}
\usepackage{courier}
\usepackage{lipsum}
\usepackage{cite}
\usepackage{amsmath,amssymb,amsfonts}
\usepackage{algorithmic}
\usepackage{graphicx}
\usepackage{textcomp}
\usepackage{xcolor}
\usepackage{url} 
\usepackage[multiple]{footmisc} 
\usepackage{subcaption} 

\usepackage{enumitem}

\usepackage{hyperref}
\usepackage{upgreek}
\usepackage{amsmath,amssymb,amsfonts}
\usepackage{float}
\usepackage{threeparttable}
\usepackage{multirow}
\usepackage{longfbox}
\usepackage{makecell}
\usepackage{longtable}
\usepackage{enumitem}
\usepackage{textcomp}
\usepackage{tcolorbox}
\newcounter{o}

\newlist{RQ}{enumerate}{1}
\setlist[RQ, 1]{label = RQ \arabic*:}

\usepackage{vcell}
\usepackage{booktabs}
\usepackage{paralist}
\usepackage{rotating}

\newcommand*\samethanks[1][\value{footnote}]{\footnotemark[#1]}

\begin{document}
%

\title{A Benchmark Study of Machine Learning Models for Online Fake News Detection}
\author[1]{Junaed Younus Khan\thanks{The authors contribute equally to this paper. Names are sorted in alphabetical order.}}
\author[1]{Md. Tawkat Islam Khondaker\samethanks}
\author[2]{Sadia Afroz}
\author[3]{Gias Uddin}
\author[1]{Anindya Iqbal}

\affil[1]{Department of Computer Science and Engineering, Bangladesh University of Engineering and Technology}
\affil[2]{International Computer Science Institute}
\affil[3]{Department of Electrical and Computer Engineering, University of Calgary}
\affil[ ]{\textit {1405051.jyk@ugrad.cse.buet.ac.bd, 1405036.mtik@ugrad.cse.buet.ac.bd, sadia@icsi.berkeley.edu, gias.uddin@ucalgary.ca, anindya@cse.buet.ac.bd}}

\renewcommand\Authands{ and }

\maketitle

\newcommand{\revision}[1]{\textcolor{black}{#1}}

\begin{abstract}
The proliferation of fake news and its propagation on social media has become a
major concern due to its ability to create devastating impacts. Different
machine learning approaches have been suggested to detect fake news. However,
most of those focused on a specific type of news (such as political) which leads
us to the question of dataset-bias of the models used. In this research, we
conducted a benchmark study to assess the performance of different applicable
machine learning approaches on three different datasets where we accumulated the largest and
most diversified one. We explored a number of advanced pre-trained language models for fake news detection along with the traditional and deep learning ones
and compared their performances from different aspects for the first time to the
best of our knowledge. We find that BERT and similar pre-trained models perform the best for fake
news detection, especially with very small dataset. Hence, these models are
significantly better option for languages with limited electronic contents,
i.e., training data. We also carried out several analysis based on the models'
performance, article's topic, article's length, and discussed different lessons
learned from them. We believe that this benchmark study will help the research
community to explore further and news sites/blogs to select the most appropriate
fake news detection method.
\end{abstract}




\section{Introduction}

Fake news can be defined as a type of yellow journalism or propaganda that
consists of deliberate misinformation or hoaxes spread via traditional print and
broadcast news media or online social media \cite{intro_3}. With the growth of
online news portals, social-networking sites, and other online media, online fake
news has become a major concern nowadays. But people are often unable to spend enough
time to cross-check references and be sure of the credibility of news. Hence, considering the scale of the users and contributors to the online media, automated detection of fake news is probably the only
way to take remedial measures, and therefore currently receiving huge attention
from the research community.


\revision{Several research works have been carried out on automated fake news detection
using both traditional machine learning and deep learning methods over the years
\cite{related_work_8,related_work_10,result_5, related_work_new_1,
related_work_new_2, related_work_new_3, related_work_new_4}. However, most of
them focused on detecting news of particular types (such as political).
Accordingly, they developed their models and designed features for specific
datasets that match their topic of interest. These approaches might suffer from
dataset bias and perform poorly on news of another topic. Hence, it is important to study
if these are sufficient for different types of news published in online media by evaluating various models on different diverse datasets and comparing their performances. However, the existing comparative studies on fake news detection methods also
focused on a specific type of dataset or explored a limited number of models. For example, Wang
built a benchmark dataset namely, Liar, and experimented some existing models on it \cite{related_work_10}. However, the length of this dataset
is not sufficient for neural network based advanced models, and some models were
found to suffer from overfitting. Gilda explored a few machine learning
approaches but did not evaluate any neural network-based model \cite{intro_4}. Recently, Gravanis et al. evaluated a number of machine
learning models on different datasets to address the issue of
dataset-bias \cite{new_benchmark}. However, they also did not explore any deep
learning based models in their study. Moreover, very
few works have been done to explore advanced pre-trained language models (e.g., BERT, ELECTRA, ELMo) for fake news detection
\cite{intro_new_1, intro_new_2} in spite of their
state-of-the-art performances in various natural language processing and text
classification tasks \cite{bert_success_1, bert_success_2, bert_success_3,
bert_success_4, bert_success_5, bert_success_6, bert_success_7}.}  

Our study fills this gap by evaluating a wide range of machine learning
approaches that include both traditional (e.g., SVM, LR, Decision Tree, Naive
Bayes, \textit{k}-NN) and deep learning (e.g., CNN, LSTM, Bi-LSTM, C-LSTM, HAN,
Conv-HAN) models on three different datasets. We have prepared a new combined
dataset containing 80k news of a great variety of topics (e.g., politics,
economy, investigation, health-care, sports, entertainment) collecting from
various sources. To the best of our knowledge, this is the largest dataset used
for fake news detection study. We also explored a variety of
pre-trained models, e.g., BERT \cite{bert_base}, RoBERTa \cite{roberta},
DistilBERT \cite{distilbert}, ELECTRA \cite{electra}, ELMo \cite{elmo_paper_1} in our comparative analysis. To the best of our
knowledge, no previous study has incorporated such advanced pre-trained models to compare
their performance with other machine learning models on fake news detection
task. In particular, we answer the following research questions.

\noindent\textbf{RQ1: How accurate are the traditional machine learning vs deep learning models to detect fake news?}  

We find that deep learning models generally outperform the traditional machine learning models, 
Among the traditional learning models, Na\"{i}ve Bayes which achieves 93\% accuracy on combined
corpus. Among the deep learning models, Bi-LSTM and
C-LSTM show great promise with 95\% accuracy on combined corpus.

\noindent\textbf{RQ2: Can the advanced pre-trained language models outperform the traditional and deep learning models?} 

We investigated pre-trained models like BERT, DistilBERT, RoBERTa, ELECTRA, and ELMo. 
Overall, these models outperform traditional and deep learning ones. For example, the pre-trained RoBERTa shows 96\% accuracy on combined corpus, which is more than the traditional and deep learning models. We also find that BERT and similar transformer-based models (BERT, DistilBERT, RoBERTa, ELECTRA) perform better than ELMo.


\noindent\textbf{RQ3: Which model performs best with small training data?} 

\revision{The superior performance of deep learning and pre-trained models we observed in our datasets could be due to large dataset sizes. However, 
the construction of a large dataset may not always be possible. We, therefore, attempted to understand whether smaller datasets can still be used to train the models without a considerable reduction in accuracy.} We see that the pre-trained models can achieve high performance with very small
training dataset compared to traditional or deep learning models. For example,
RoBERTa achieved over 90\% accuracy with only 500 training data used for
fine-tuning while traditional and deep learning models fail to achieve even 80\%
accuracy with such small dataset (see Figure
\ref{rq4_comparison_of_best_models_on_small_part_fake_or_real}). In contrast, the best performing traditional learning model Na\"{i}ve Bayes only achieved 65\% accuracy with a sample size of 500 training set.
Therefore, our finding can be useful for
electronic-resource-limited languages where fake news dataset collections are
likely to be small in size. In such cases, based on our observations, pre-trained models are the best option to achieve
quality performance for these languages. Note that different languages such as
Dutch, Italian, Arabic, Bangla, etc. have pre-trained BERT models
\cite{dutch_bert, italian_bert, arabic_bert} that can be fine-tuned with small
fake news dataset to develop detection tool.

\noindent\textbf{Replication Package} with code and data is shared online at \url{https://github.com/JunaedYounusKhan51/FakeNewsDetection}.

\noindent\textbf{Paper Organizations.} 
The rest of this paper is structured as follows. In Section
\ref{sec:related_work}, we compare related research works. In Section \ref{sec:materials_and_methods},
we describe our study setup by introducing the datasets, the
features, and the models we used in our experiments. Section
\ref{sec:results} presents the performance of different models on
three datasets and answer three research questions. Section
\ref{sec:discussion} compares the performance and analyzes the misclassified cases. We conclude in Section \ref{sec:conclusions}. 


\section{Related Work}
\label{sec:related_work}



Related work can broadly be divided into the following categories:
\begin{inparaenum}[(1)]
\item Exploratory analysis of the characteristics of fake news,  
\item Traditional machine learning based detection,
\item Deep learning based detection,
\item Advanced language model based detection, and
\item Benchmark studies.
\end{inparaenum}  

\subsection{Exploratory analysis of the characteristics of fake news}
Several research works have been done over the years on the characteristics of
fake news and its' detection. Conroy et al. mentioned
three types of fake news: Serious Fabrications, Large-Scale Hoaxes, and Humorous
Fakes \cite{related_work_1}. They have termed fake news as a news article that
is intentionally and verifiably false and could mislead readers
\cite{related_work_2}. This narrow definition is useful in the sense that it can eliminate the ambiguity between fake news and other related concepts,
e.g., hoaxes, and satires.

\subsection{Traditional machine learning based detection}
Different traditional machine learning based approaches have
been proposed for the automatic detection of fake news. 
In \cite{related_work_3}, the authors proposed to use linguistic-based features
such as total words, characters per word, frequencies of large words,
frequencies of phrases, i.e., “n-grams” and bag-of-words approaches
\cite{related_work_4}, parts-of-speech (POS) tagging for fake news detection.

Conroy et al. argued that simple content-related n-grams and part-of-speech
(POS) tagging had been proven insufficient for the classification task
\cite{related_work_5}. Rather, they suggested Deep Syntax analysis using
Probabilistic Context-Free Grammars (PCFG) following another work by Feng et al.
\cite{related_work_6} to distinguish rule categories (i.e., lexicalized,
non-lexicalized, parent nodes, etc.) for deception detection with 85-91\%
accuracy. However, Shlok Gilda reported that while bi-gram TF-IDF yielded highly
effective models for detecting fake news, the PCFG features had little to add to
the models’ efficacy \cite{intro_4}.

Many research works also suggested the use of sentiment analysis for deception
detection as some correlation might be found between the sentiment of the news
article and its type. Reference \cite{related_work_8} proposed expanding the
possibilities of word-level analysis by measuring the utility of features like
part of speech frequency, and semantic categories such as generalizing terms,
positive and negative polarity (sentiment analysis).

Cliche described the detection of sarcasm on twitter using n-grams, words
learned from tweets specifically tagged as sarcastic \cite{related_work_9}. His
work also included the use of sentiment analysis as well as identification of
topics (words that are often grouped together in tweets) to improve prediction
accuracy.

\subsection{Deep learning based detection}
Several research works used deep learning models to detect fake news. Wang et
al. built a hybrid convolutional neural network model that outperforms other
traditional machine learning models \cite{related_work_10}. Rashkin et al.
performed an extensive analysis of linguistic features and showed promising
result with LSTM \cite{method_5}. Singhania et al. proposed a three-level
hierarchical attention network, one each for words, sentences, and the headline
of a news article \cite{old_new_related_work_1}. Ruchansky et al. created the
CSI model where they have captured text, the response of an article, and the
source characteristics based on users’ behaviour \cite{old_new_related_work_2}.

Among the recent works, Shu et al. argued that a critical aspect of fake news
detection is the explainability of such detection in \cite{related_work_new_1}.
The authors developed a sentence-comment co-attention sub-network to exploit
both news contents and user comments. In this way, the authors focused on
jointly capturing explainable check-worthy sentences and user comments for fake
news detection. In the work \cite{related_work_new_2}, the authors developed a
multimodal variational auto-encoder by using a bi-modal variational auto-encoder
coupled with a binary classifier for the task of fake news detection. The
authors claimed that this end-to-end network utilizes the multimodal
representations obtained from the bi-modal variational auto-encoder to classify
posts as fake or not. Zhou et al. focused on studying the patterns of spreading
of fake news in social networks, and the relationships among the spreaders
\cite{related_work_new_3}. Hamdi et al. proposed a hybrid approach to detect
misinformation in Twitter \cite{related_work_new_5}. The authors extracted user
characteristics using node2vec to verify the credibility of the contents.

\begin{table}
\caption{Comparison between our benchmark study and prior benchmark studies}
\centering
\resizebox{\columnwidth}{!}{%
\revision{
\scalebox{0.5}
{
\begin{tabular}{l|l|l|l} 
\toprule
\multicolumn{1}{c|}{\textbf{Theme}}                                                 & \multicolumn{1}{l|}{\textbf{Prior Benchmark Study} }                                                                                                                                      & \multicolumn{1}{l|}{\textbf{Limitations} }                                                                                                                                                                     & \multicolumn{1}{l}{\textbf{Our Benchmark Study}}                                                                                                                   \\ 
\midrule
\multirow{3}{*}{\begin{tabular}[c]{@{}l@{}}Experiment\\and\\Result\end{tabular}}     & \begin{tabular}[c]{@{}l@{}}Bondielli et al. surveyed\\the different approaches\\to automatic detection of\\fake news in the recent\\literature~\\\cite{bondielli_benchmark_2019}.\end{tabular} & \multirow{3}{*}{\begin{tabular}[c]{@{}l@{}}They did not run any\\experiments and\\did not report any\\results. \end{tabular}}                                                                                  & \multirow{3}{*}{\begin{tabular}[c]{@{}l@{}}We experimented with \\all the models and\\analyzed their\\performances. \end{tabular}}                                  \\ 
\cmidrule{2-2}
                                                                                     & \begin{tabular}[c]{@{}l@{}}Dwivedi et al. presented\\a literature survey on\\various fake news\\detection methods~\\\cite{dwivedi_benchmark_2020}.\end{tabular}                                &                                                                                                                                                                                                                &                                                                                                                                                                     \\ 
\cmidrule{2-2}
                                                                                     & \begin{tabular}[c]{@{}l@{}}Zhang et al. presented an\\overview of the existing \\datasets and fake news \\detection approaches\\\cite{fakenewsoverview2020zhang}.\end{tabular}                   &                                                                                                                                                                                                                &                                                                                                                                                                     \\ 
\midrule
\begin{tabular}[c]{@{}l@{}}Dataset\\Length\\and\\Diversity\end{tabular}              & \begin{tabular}[c]{@{}l@{}}Wang experimented with\\some existing models\\on their benchmark\\dataset namely, Liar~\\\cite{related_work_10}.\end{tabular}                                       & \begin{tabular}[c]{@{}l@{}}They evaluated the~\\models only on one\\dataset. Moreover,\\the length of the\\dataset was not\\sufficient, and some\\models were found\\to suffer from\\overfitting.\end{tabular} & \begin{tabular}[c]{@{}l@{}}We evaluated all the\\methods on three\\different and diverse\\datasets.\end{tabular}                                                    \\ 
\midrule
\multirow{3}{*}{\begin{tabular}[c]{@{}l@{}}Range of\\Models~\\Explored\end{tabular}} & \begin{tabular}[c]{@{}l@{}}Gilda explored a few\\machine learning\\approaches for fake news\\detection\\\cite{intro_4}. \end{tabular}                                                           & \multirow{2}{*}{\begin{tabular}[c]{@{}l@{}}They did not evaluate\\any deep learning\\based model. \end{tabular}}                                                                                               & \multirow{3}{*}{\begin{tabular}[c]{@{}l@{}}We explored deep\\learning and advanced \\pre-trained language\\models along with\\traditional ones.~ ~ ~\end{tabular}}  \\ 
\cmidrule{2-2}
                                                                                     & \begin{tabular}[c]{@{}l@{}}Gravanis et al. evaluated\\a number of machine\\learning models on\\different datasets \\\cite{new_benchmark}.\end{tabular}                                          &                                                                                                                                                                                                                &                                                                                                                                                                     \\ 
\cmidrule{2-3}
                                                                                     & \begin{tabular}[c]{@{}l@{}}Oshikawa et al. compared\\various existing methods\\for fake news detection\\on different datasets\\\cite{oshikawa_benchmark_2018}.\end{tabular}                     & \begin{tabular}[c]{@{}l@{}}They did not explore\\advanced language\\models such as BERT,\\ELECTRA, ELMo, etc \end{tabular}                                                                                        &                                                                                                                                                                     \\
\bottomrule
\end{tabular}}%
}
}
\label{table:comparison_related_work}
\end{table}

\subsection{Advanced language model based detection}
Currently, Advanced pre-trained language models (i.e., BERT, ELECTRA, ELMo) are receiving great attention for several natural
language tasks including text classification \cite{bert_success_1,
bert_success_2, bert_success_3, bert_success_4, bert_success_5, bert_success_6,
bert_success_7}. However, only a few studies have explored them for fake news
detection. For example, Jwa et al. detected fake news by analyzing the
relationship between the headline and the body text of news 
\cite{intro_new_1}. The authors claimed that the deep-contextualizing nature of
BERT improves F-score by 0.14 over the previous state-of-the-art models. Kula et
al. presented a hybrid architecture connecting BERT with RNN to tackle the
impact of fake news \cite{intro_new_2}. Lee et al. worked on hyperpartisan
dataset and leveraged BERT on semi-supervised pseudo-label dataset
\cite{related_work_new_6}.

\subsection{Benchmark studies}
While most of the existing researches have focused on defining the
types of fake news and suggesting different approaches to detect them, very few
studies are carried out to compare such approaches independently on different
datasets. Among the categories, the 
benchmark-based studies are the most similar to our study. Table \ref{table:comparison_related_work} 
compares our work with the previous benchmark-based studies along three themes: \begin{inparaenum}[(1)]
\item experimental setup and results, 
\item dataset length and diversity, and 
\item range of models explored.
\end{inparaenum} We discuss the related work below. 

Wang et al. compared the performance of SVM, LR, Bi-LSTM, and CNN
models on their proposed dataset “LIAR” \cite{related_work_10}. Oshikawa et al. compared various machine learning models (e.g., SVM, CNN, LSTM) for fake news detection on different datasets \cite{oshikawa_benchmark_2018}. 
Gravanis et al. compared several traditional machine learning models (i.e.,
\textit{k}-NN, Decision Tree, Naive Bayes, SVM, AdaBoost, Bagging) for fake news
detection on different datasets \cite{new_benchmark}. Dwivedi et al. presented
a literature survey on various fake news detection methods \cite{dwivedi_benchmark_2020}. Zhang et al. presented a comprehensive overview of the existing datasets and approaches proposed for fake news
detection in previous literature \cite{fakenewsoverview2020zhang}.

\revision{In summary, these few existing comparative studies lack in terms of
the range of evaluated models and the diversity of the used datasets. Moreover,
a complete exploration of the advanced pre-trained language models for fake news detection and
comparison among them and with other models (i.e., traditional and deep learning) were missing in previous works. The benchmark study presented in this paper is focused on dealing with
the above issues. We extend the state-of-the-art research in fake news detection by offering a comprehensive an in-depth study of 19 models (eight traditional shallow learning models, six traditional deep learning models, and five advanced pre-trained language models).}

\section{Study Setup}
\label{sec:materials_and_methods}
In this section, we first introduce the datasets used in our study and discuss how we preprocess those (Section \ref{sec:studied-datasets}). 
Then we discuss different features that we used in our models in Section \ref{sec:studied-features}. Finally, we discuss the traditional learning, deep learning and pre-trained models 
that we investigated in our study (Section \ref{sec:studied-models}). Finally, we discuss the performance metrics we used to evaluate the models and the train and 
test data settings in Section \ref{sec:evaluation-metrics}. 
\subsection{Studied Datasets}\label{sec:studied-datasets}
In this comparative study, we make use of three following datasets. Table \ref{dataset_table} shows the detailed statistics of them. 
We describe the datasets below.
\begin{table*}[t]
\caption{Properties of Datasets}
\begin{center}
\resizebox{\columnwidth}{!}
{
\begin{tabular}{lrrrcl}
\toprule
\textbf{Dataset} & \textbf{\begin{tabular}[c]{@{}r@{}}\#Total \\   Data\end{tabular}} & \textbf{\begin{tabular}[c]{@{}l@{}}\#Fake \\   
News\end{tabular}} & \textbf{\begin{tabular}[c]{@{}r@{}}\#Real \\   News\end{tabular}} & \textbf{\begin{tabular}[c]{@{}c@{}}Avg. Length of \\ News Articles \\ (in words)\end{tabular}} & \textbf{Topic(s)} \\ 
\midrule
LIAR & 12791 & 5657 & 7134 & 18 & Politics \\ \midrule
Fake or Real News & 6335 & 3164 & 3171 & 765 & \begin{tabular}[c]{@{}l@{}}Politics \\ (2016 USA election)\end{tabular} \\ \midrule 
Combined Corpus & 79548 & 38859 & 40689 & 644 & \begin{tabular}[c]{@{}l@{}}Politics, Economy, \\ Investigation, Health, \\ Sports, Entertainment\end{tabular} \\ 
\bottomrule
\end{tabular}
}
\label{dataset_table}
\end{center}
\end{table*}

\subsubsection{Liar}
Liar\footnote{\url{https://www.cs.ucsb.edu/~william/data/liar_dataset.zip}} is a publicly available dataset that has been used in \cite{related_work_10}. It includes 12.8K human-labeled short statements from POLITIFACT.COM. 
It comprises six labels of truthfulness ratings: pants-fire, false, barely-true, half-true, mostly-true, and true. In our work, we try to differentiate real news from all types of hoax, propaganda, satire, and misleading news. Hence, we mainly focus on classifying news as real and fake. For the binary classification of news, we transform these labels into two labels. Pants-fire, false, barely-true are contemplated as fake and half-true, mostly-true, and true are as true. Our converted dataset contains 56\% true and 44\% fake statements. This dataset mostly deals with political issues that include statements of democrats and republicans, as well as a significant amount of posts from online social media. The dataset provides some additional meta-data like the subject, speaker, job, state, party, context, history. However, in the real-life scenario, we may not have this meta-data always available. Therefore, we experiment on the texts of the dataset using textual features.

\subsubsection{Fake or Real News}
Fake or real news dataset 
is developed by George McIntire. The fake news portion of this dataset was collected from Kaggle fake news dataset\footnote{\url{https://www.kaggle.com/mrisdal/fake-news}} comprising news of the 2016 USA election cycle. The real news portion was collected from media organizations such as the New York Times, WSJ, Bloomberg, NPR, and the Guardian for the duration of 2015 or 2016. The GitHub repository of the dataset includes around 6.3k news with an equal allocation of fake and real news, and half of the corpus comes from political news.

\subsubsection{Combined Corpus}
\begin{figure}[htbp]
\centerline{\includegraphics[height=8cm,keepaspectratio]{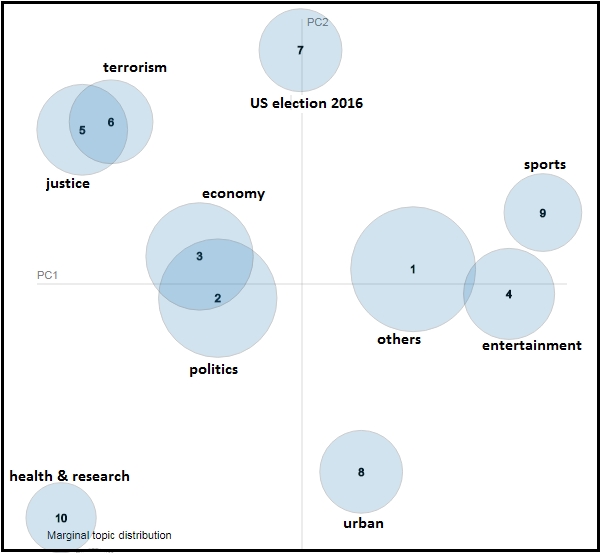}}
\caption{Inter-topic distance map of Combined Corpus.}
\label{dataset_fig_1_LDA_Topic}
\end{figure}

Apart from the other two datasets, we have built a combined corpus that contains around 80k news among which 51\% are real, and 49\% are fake. One important property of this corpus is that it incorporates a wide range of topics including national and international politics, economy, investigation, health-care, sports, entertainment, and others. \revision{To demonstrate the topic diversity, we show the inter-topic distances\footnote{generated using pyLDAvis: \url{https://pyldavis.readthedocs.io/}} of our combined corpus using LDA-based (Latent Dirichlet Allocation) topic modeling \cite{combined_corpus_new_1} in Figure~\ref{dataset_fig_1_LDA_Topic}.} Based on the empirical analysis of inter-topic distances, we divided the dataset into ten clusters (circles) where each cluster represents a topic. \revision{The coordinates of each topic cluster (circle) were measured following the MDS (Multidimensional Scaling) algorithm \cite{LDA_MDS}. X-axis (PC1) and Y-axis (PC2) maintained an aspect ratio to 1 to preserve the MDS distances. We used Jensen-Shannon divergence \cite{LDA_jensen_shannon} to compute distances between topics.} \revision{The area of a cluster was calculated by the portion of tokens that respective topic generated compared to the total tokens in the corpus.} We named the topic of a cluster based on the most relevant terms representing that cluster. The most relevant terms were determined on the basis of frequency. For example, the most relevant (i.e., most frequent) terms for cluster-7 are `Trump', `Clinton', `Election', `Campaign', etc (Figure~\ref{dataset_fig_1_LDA_Topic}). Hence, the news of this cluster represents the 2016 US election. On the other hand, the most relevant terms for cluster-3 are 'Bank', 'Job', 'Financial', 'Tax', 'Market', etc. Thus, this cluster is related to the Economy. Additionally, overlapping of clusters (e.g., Economy and Politics) indicates shared relevant words (e.g., `Government', `People') between them. We have collected news from several sources of the same time domain mostly from 2015 to 2017 \footnote{\url{https://homes.cs.washington.edu/~hrashkin/factcheck.html}}\textsuperscript{,}\footnote{\url{https://github.com/suryattheja/Fake-news-detection}}\textsuperscript{,}\footnote{\url{https://www.kaggle.com/snapcrack/all-the-news}}. Multiple types of fake news such as hoax, satire, and propaganda have come from The Onion, Borowitz Report, Clickhole, American News, DC Gazette, Natural News, and  Activist Report. We have collected the real news from the trusted sources like the New York Times, Breitbart, CNN, Business Insider, the Atlantic, Fox News, Talking Points Memo, Buzzfeed News, National Review, New York Post, the Guardian, NPR, Gigaword News,  Reuters, Vox, and the Washington Post.  

\subsubsection{Data Preprocessing}  \label{sec:data-preprocessing}
Before feeding into the models, raw texts of news required some preprocessing. We first eliminated unnecessary IP and URL addresses from our texts. The next step was to remove stop words. After that, we cleaned our corpus by correcting the spelling of words. We split every text by white-space and remove suffices from words by stemming them. Finally, we rejoined the word tokens by white-space to present our clean text corpus which had been tokenized later for feeding into the models.


\subsection{Studied Features} \label{sec:studied-features}
We used lexical and sentiment features, n-gram, and Empath generated features for traditional machine learning models, and pre-trained word embedding for deep learning models. 
%
%
\subsubsection{Lexical and Sentiment Features}
Several studies have proposed to use lexical and sentiment features for fake news detection \cite{related_work_8, related_work_3, method_5}. For lexical features, we used word count, average word length, article length, count of numbers, count of parts of speech, and count of exclamation mark. We calculated the sentiment (i.e., positive and negative polarity) of every article and used them as sentiment features.

%
%
\subsubsection{n-gram Feature}
Word-based n-gram was used to represent the context of the document and generate features to classify the document as fake and real \cite{method_5, background_18,method_4,method_6,method_7,method_9}. We used both uni-gram and bi-gram features in this benchmark and evaluated their effectiveness.

%
%
\subsubsection{Empath Generated Features}
Empath is a tool that can generate lexical categories from a given text using a small set of seed terms\cite{background_24}. Using Empath, we calculated these categories (e.g., violence, crime, pride, sympathy, deception, war) for every news data and used them as features to identify key information in a news article. Since it has been used in literature for understanding deception in review systems \cite{background_24}, we feel motivated to investigate their contribution in this context.

%
%
\subsubsection{Pre-trained Word Embedding}
For neural network models, word embeddings were initialized with 100-dimensional pre-trained embeddings from GloVe \cite{method_10}. GloVe is an unsupervised learning algorithm for obtaining vector representations for words. It was trained on a dataset of one billion tokens (words) with a vocabulary of 400 thousand words.


\subsection{Studied Models} \label{sec:studied-models}
We experimented various traditional, deep learning and pre-trained language models in this work. Here, we describe all the models that we studied. 

\subsubsection{Traditional Machine Learning Models} 
We built our first three models using SVM (Support Vector Machine), LR (Logistic
Regression), and Decision Tree with the lexical and sentiment features. Among
the four main variants of the SVM kernel, we used the linear one. We also
evaluated ensemble learning method like AdaBoost combining 30 decision trees
with lexical and sentiment features. Next, we explored the Multinomial Naive
Bayes classifier with the n-gram features. We used the Empath generated features
with \textit{k}-NN (\textit{k}-Nearest Neighbors) classifier. We use the
square-root of the total training data size as \textit{k} as suggested by Lall
and Sharma \cite{knn_opitmal_k}. Hence, the value of \textit{k} was chosen to
be 70, 90, and 250 for Liar, Fake or Real, and Combined Corpus respectively.
%
%

%
%

%
%

\subsubsection{Deep Learning Models}
In this study, we have evaluated six deep learning models for fake news
detection including CNN, LSTM, Bi-LSTM, C-LSTM, HAN, and Convolutional HAN. The
models are described below with their experimental setups.

\begin{inparaenum}[(1)]
\item\textbf{\textit{CNN:}} \revision{One dimensional convolutional neural network can extract features and classify texts after transforming words in the sentence corpus into vectors \cite{new_background_1}.}The one-dimensional convolutional model was initialized with 100-dimensional pre-trained GloVe embeddings. It contained 128 filters of filter size 3 and a max pooling layer of pool size 2 is selected. A dropout probability of 0.8 was preserved which was expunged for Combined Corpus. The model was compiled with ADAM optimizer with a learning rate of 0.001 to minimize binary cross-entropy loss. A sigmoid activation function was used for the final output layer. A batch size of 64 and 512 was used for training the datasets over 10 epochs.

\item\textbf{\textit{LSTM:}} Our LSTM model was pre-trained with 100-dimensional
GloVe embeddings.  The output dimension and time steps were set to 300. ADAM
optimizer with learning rate 0.001 was applied to minimize binary cross-entropy
loss. Sigmoid was the activation function for the final output layer.
The model was trained over 10 epochs with batch size 64 and 512.

\item\textbf{\textit{Bi-LSTM:}} Usually, news that is deemed as fake is not fully comprised of false information, rather it is blended with true information. To detect the anomaly in a certain part of the news, we need to examine it both with previous and next events of action. We constructed a Bi-LSTM model to perform this task. Bi-LSTM was initialized with 100-dimensional pre-trained GloVe embeddings. The output dimension of 100 and time steps of 300 was applied. ADAM optimizer with a learning rate of 0.001 was used to minimize binary cross-entropy loss. The training batch size was set to 128 and loss over each epoch was observed with a callback. The learning rate was reduced by a factor of 0.1. We also used an early stop to monitor validation accuracy to check whether the accuracy was deteriorating for 5 epochs. The loss of the binary cross-entropy of the model was minimized by ADAM with a learning rate of 0.0001.

\item\textbf{\textit{C-LSTM:}} The C-LSTM based model contained one convolutional layer and one LSTM layer. We used 128 filters with filter size 3 on top of which a max pooling layer of pool size 2 was set. We fed it to our LSTM architecture with 100 output dimensions and dropout 0.2. Finally, we used sigmoid as the activation function of our output layer.

\item\textbf{\textit{HAN:}} We used a hierarchical attention network consisting of two attention mechanisms for word-level and sentence-level encoding. Before training, we set the maximum number of sentences in a news article as 20 and the maximum number of words in a sentence as 100. In both level encoding, a bidirectional GRU with output dimension 100 was fed to our customized attention layer. We used word encoder as input to our sentence encoder time-distributed layer. We optimized our model with ADAM that learned at a rate of 0.001.

\item\textbf{\textit{Convolutional HAN:}} In order to extract high-level features of the input, we incorporated a one-dimensional convolutional layer before each bidirectional GRU layer in HAN. This layer selected features of each tri-gram from the news article before feeding it to the attention layer.
\end{inparaenum}

\subsubsection{Advanced Language Models}   
Here, we first discuss the advanced language models that we used in this study and then describe their experimental setup.

\begin{inparaenum}[(1)]
\item \textbf{\textit{BERT:}} BERT (Bidirectional Encoder Representations from Transformers) is a pre-trained model which was designed to learn contextual word representations of unlabeled texts \cite{bert_base}. Among the two versions of BERT (i.e., BERT-Base and BERT-Large) proposed originally, we used BERT-Base for this study considering the huge time and memory requirements of the BERT-Large model. The BERT-Base model has 12 layers (transformer blocks) with 12 attention heads and 110 million parameters.

\item \textbf{\textit{RoBERTa:}} RoBERTa (Robustly optimized BERT approach), originally suggested in \cite{roberta}, is the second pre-trained model that we experimented. It achieves better performance than original BERT models by using larger mini-batch sizes to train the model for a longer time over more data. It also removes the NSP loss in BERT and trains on longer sequences. Moreover, it dynamically changes the masking pattern applied to the training data.

\item\textbf{\textit{DistilBERT:}} DistilBERT \cite{distilbert} is a smaller, faster, cheaper, and lighter version of original BERT which has 40\% fewer parameters than the BERT-Base model. Though original BERT models perform better, DistilBERT is more appropriate for production-level usage due to its' low resource requirements. Considering potential users of non-profit blogs and online media, we think low-resource models have a good appeal. Hence, this is worth investigating.

\revision{
\item\textbf{\textit{ELECTRA:}} ELECTRA (Efficiently Learning an Encoder that Classifies Token Replacements Accurately) \cite{electra} is a transformer model for self-supervised language representation learning. This model pre-trained with the use of another (small) masked language model. First, a language model takes an input text and randomly masked the text with generated input token. Then, ELECTRA models are trained to distinguish "real" input tokens vs "fake" input tokens generated by the former language model. At small scale, ELECTRA can achieve strong results even when trained on a single GPU. 
}

\revision{
\item\textbf{\textit{ELMo:}}
ELMo (Embeddings from Language Models) is a contextualized word representation learned from a deep bidirectional language model that is trained on a large text corpus \cite{elmo_paper_1}. We used the original pre-trained ELMo model proposed by the authors that has 2 bi-LSTM layers and 93.6 million parameters.
}
\end{inparaenum}

\textbf{\textit{Experimental Setup of Advanced Language Models:}}

\begin{figure}[t]
\centerline{\includegraphics[]{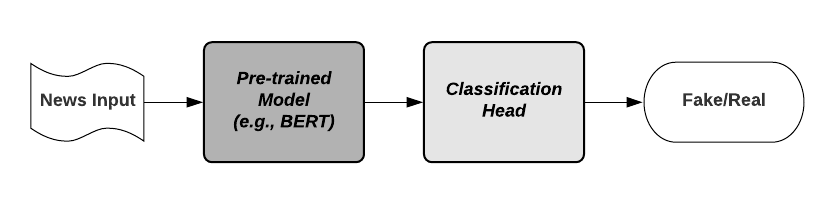}}
\caption{Fine-tuning of pre-trained language models.}
\label{fig:finetuning}
\end{figure}

\revision{We appended a classification head composed of a single linear layer on the top of the pre-trained advanced language models. The architecture of the classifier head is kept simple to focus on what
information can readily be extracted from these pre-trained models. We used the respective pre-trained embeddings of the corresponding models (e.g., BERT embeddings, ELECTRA embeddings, ELMo embeddings) as the input of the classification heads and fine-tuned them for the fake news detection task (Figure \ref{fig:finetuning}).} We trained them on all the datasets for 10 epochs with a mini-batch size of 32. We applied early stop to prevent our models from overfitting \cite{early_stop_1}. Validation loss was considered as the metric of the early stopping while delta is set to zero \cite{early_stop_2}. We set the maximum sequence length of the input data to 300. For the Combined Corpus dataset, we configured the gradient accumulation steps as 2 due to the large dataset size. We used AdamW optimizer \cite{adam_w} with the learning rate set to 4e-5, ß1 to 0.9, ß2 to 0.999, and epsilon to 1e-8 \cite{bert_base, bert_fine_tuning}. Finally, we used binary cross-entropy to calculate the loss \cite{binary_cross}. We performed the experiments on NVIDIA Tesla T4 GPU provided by Google Colab.

\subsection{Evaluation Metrics} \label{sec:evaluation-metrics}
We created a standard training and test set for each of the three datasets by
splitting it in an 80:20 ratio so that different models can be evaluated on the
same ground. For the first two datasets (i.e., Liar, Fake or Real), we did the
split randomly as they only contain one type of news. On the other hand, as the
Combined Corpus covers a wide variety of topics, we took 80\% (20\%) data from
each topic and include them in train (test) set to maintain a balanced
distribution of every topic in training and test data.

We report the performance of each model in terms of accuracy, precision, recall, and F1-score. For precision, recall, and F1-score, we considered the macro-average of both class.

In our experiment, we considered real news as `positive class', and fake news as
`negative class'. Hence, True Positive (TP) means the news is actually real, and
also predicted as real while False Positive (FP) indicates that the news is
actually false, but predicted as real. True Negative (TN) and False Negative
(FN) imply accordingly. Accuracy is the number of correctly predicted instances out of all instances.
\begin{equation}
Accuracy (A) = \frac{TP+TN}{TP+FN+TN+FP}
\end{equation}

Precision is the ratio between the number of correctly predicted instances and all the predicted instances for a given class. For real and fake classes, we presented this metric as P(R) and P(F) respectively. Hence, the macro-average precision, P will be the average of P(R) and P(F).
\begin{equation}
P(R) = \frac{TP} {TP + FP},~P(F) = \frac{TN} {TN + FN},~P = \frac{P(R) + P(F)} {2}
\end{equation}

Recall represents the ratio of the number of correctly predicted instances and all instances belonging to a given class. For real and fake classes, we presented this metric as R(R) and R(F) respectively. Hence, the macro-average recall, R will be the average of R(R) and R(F).
\begin{equation}
R(R) = \frac{TP} {TP + FN},~R(F) = \frac{TN} {TN + FP},~R = \frac{R(R) + R(F)} {2}
\end{equation}

F1-score is the harmonic mean of the precision and recall.
\begin{equation}
F1 = \frac{2\cdot P\cdot R}{P + R}
\end{equation}


\section{Study Results}
\label{sec:results}
\noindent In this section, we answer three research questions:
\begin{RQ}[label=\textbf{RQ\arabic{*}}., leftmargin=35pt]
  \item How accurate are the traditional and deep learning models to detect fake news in our datasets?
  \item Can the advanced pre-trained language models outperform the traditional and deep learning models?
  \item Which model performs best with small training data?
\end{RQ}
Previous studies on fake news detection mainly focused on traditional
machine learning models. Therefore, it is important to compare their performance with
the deep learning models. We address this concern in RQ1. In particular, the goal of RQ1 
is to compare the performance of different traditional
machine learning models (e.g., SVM, Naive Bayes, Decision Tree) and deep
learning models (e.g., CNN, LSTM, Bi-LSTM) on fake news detection.
Considering the great success of pre-trained advanced language models on various text classification tasks, it is important to investigate how these models perform on fake news detection compared to the
traditional and deep learning models. The answers to RQ2 will offer insights into 
whether and how the pre-trained advanced language models are useful to detect fake news. 
A common issue for any supervised
learning problem is the limitation of labeled data. Intuitively, the more performance we can get with less amount of labeled data, the easier it would be to investigate and 
develop machine learning models to facilitate fake news detection. Therefore, as 
part of RQ3, we investigate the performance of the models we used in our study on smaller samples of 
our datasets.


\subsection{How accurate are the traditional and deep learning models to detect fake news in our datasets? (RQ1)}


\begin{table*}[t]
\caption{Performance of Traditional Machine Learning Models}
\resizebox{\columnwidth}{!}{%
\begin{tabular}{llrrrr|rrrr|rrrr}
 &  & \multicolumn{12}{c}{\textbf{Datasets}} \\ \cmidrule{3-14} 
 &  & \multicolumn{4}{c}{\textit{\textbf{Liar}}} & \multicolumn{4}{c}{\textit{\textbf{Fake or Real News}}} & \multicolumn{4}{c}{\textit{\textbf{Combined Corpus}}} \\ \midrule
{\textbf{Model}} & \textbf{Feature} & \bf{A} & \bf{P} & \bf{R} & \bf{F1} & \bf{A} & \bf{P} & \bf{R} & \bf{F1} & \bf{A} & \bf{P} & \bf{R} & \bf{F1} \\ \midrule
{SVM} & Lexical & .56 & .56 & .56 & .48 & .67 & .67 & .67 & .67 & .71 & .78 & .71 & .72 \\ \midrule
{SVM} & \begin{tabular}[c]{@{}l@{}}Lexical\\ +Sentiment\end{tabular} & .56 & .57 & .56 & .48 & .66 & .66 & .66 & .66 & .71 & .77 & .71 & .72 \\ \midrule
{LR} & \begin{tabular}[c]{@{}l@{}}Lexical \\ +Sentiment\end{tabular} & 0.56 & .56 & .56 & .51 & .67 & .67 & .67 & .67 & .76 & .79 & .76 & .77 \\ \midrule
{\begin{tabular}[c]{@{}l@{}}Decision \\ Tree\end{tabular}} & \begin{tabular}[c]{@{}l@{}}Lexical \\ +Sentiment\end{tabular} & .51 & .51 & .51 & .51 & .65 & .65 & .65 & .65 & .67 & .71 & .69 & .7 \\ \midrule
{AdaBoost} & \begin{tabular}[c]{@{}l@{}}Lexical\\ +Sentiment\end{tabular} & .56 & .56 & .56 & .54 & .72 & .72 & .72 & .72 & .73 & .74 & .73 & .74 \\ \midrule
{\begin{tabular}[c]{@{}c@{}}Naive \\ Bayes\end{tabular}} & \begin{tabular}[c]{@{}l@{}}Unigram\\ (TF-IDF)\end{tabular} & .60 & .60 & .60 & .57 & .82 & .82 & .82 & .82 & .91 & .91 & .91 & .91 \\\midrule
{\begin{tabular}[c]{@{}c@{}}Naive \\ Bayes\end{tabular}} & \begin{tabular}[c]{@{}l@{}}Bigram \\ (TF-IDF)\end{tabular} & \textbf{.60} & \textbf{.59} & \textbf{.60} &\textbf{ .59} &\textbf{ .86} & \textbf{.86} & \textbf{.86} & \textbf{.86} & \textbf{.93} & \textbf{.93} & \textbf{.93} & \textbf{.93} \\ \midrule
{\textit{k}-NN} & \begin{tabular}[c]{@{}l@{}}Empath \\ Features\end{tabular} & .54 & .54 & .54 & .54 & .71 & .72 & .71 & .71 & .71 & .70 & .70 & .70 \\ 
\bottomrule
\end{tabular}%
}
\label{result_table_1}
\end{table*}

In Table \ref{result_table_1}, we report the performances of various traditional machine learning models in detecting fake news. We observe that among the traditional machine learning models, Naive Bayes with n-gram features performs the best with 93\% accuracy on our Combined Corpus. We also find that the addition of sentiment features with lexical features does not improve the performance considerably. For lexical and sentiment features, SVM and LR models perform better than other traditional machine learning models as suggested by most of the prior studies \cite{related_work_8,related_work_10,result_5,result_3,background_12}. On the other hand, Empath generated features do not show promising performance for fake news detection, although they had been used earlier for understanding deception in review systems \cite{background_24}.


\begin{table*}[ht]
\caption{Performance of Deep Learning Models (using Glove word embedding as feature)}
\resizebox{\columnwidth}{!}{
\begin{tabular}{lrrrr|rrrr|rrrr}
   & \multicolumn{12}{c}{\textbf{Datasets}} \\ 
 \cmidrule{2-13}
   & \multicolumn{4}{c}{\textit{\textbf{Liar}}} & \multicolumn{4}{c}{\textit{\textbf{Fake or Real News}}} & \multicolumn{4}{c}{\textit{\textbf{Combined Corpus}}} \\ 
   \midrule
{\textbf{Model}} &  \bf{A} & \bf{P} & \bf{R} & \bf{F1} &  \bf{A} & \bf{P} & \bf{R} & \bf{F1} &  \bf{A} & \bf{P} & \bf{R} & \bf{F1} \\ 
\midrule
{CNN}  & .58 & .58 & .58 & .58 & .86 & .86 & .86 & .86 & .93 & .93 & .93 & .93 \\ 
\midrule 
{LSTM}  & .54 & .29 & .54 & .38 & .76 & .78 & .76 & .76 & .93 & .94 & .93 & .93 \\ 
\midrule
{Bi-LSTM}  & .58 & .58 & .58 & .57 & .85 & .86 & .85 & .85 & \textbf{.95} & \textbf{.95} & \textbf{.95} & \textbf{.95} \\ \midrule 
{C-LSTM}  & .54 & .29 & .54 & .38 & .86 & .87 & .86 & .86 & \textbf{.95} & \textbf{.95} & \textbf{.95} & \textbf{.95} \\ \midrule 
{HAN} &  .57 & .57 & .57 & .56 & \textbf{.87} & \textbf{.87} & \textbf{.87} & \textbf{.87} & .92 & .92 & .92 & .92 \\ \midrule 
Conv-HAN  & \textbf{.59} & \textbf{.59} & \textbf{.59} & \textbf{.59} & .86 & .86 & .86 & .86 & .92 & .92 & .92 & .92 \\ 
\bottomrule
\end{tabular}%
}
\label{result_table_2}
\end{table*}

In Table \ref{result_table_2}, we report the performances of different deep learning models. The baseline CNN model is considered as the best model for Liar in \cite{related_work_10}, but we find it to be the second-best among all the models. LSTM-based models are most vulnerable to overfitting for this dataset which is reflected by its performance. Although Bi-LSTM is also a victim of overfitting on the Liar dataset as mentioned in \cite{related_work_10}, we find it to be the third-best neural network-based model according to its performance on the dataset. The models successfully used for text classification like C-LSTM, HAN hardly surmount the overfitting problem for the Liar dataset. Our hybrid Conv-HAN model exhibits the best performance among the neural models for the Liar dataset with 0.59 accuracy and 0.59 F1-score. LSTM-based models show an improvement on the Fake or Real dataset whereas CNN and Conv-HAN continue their impressive performance. LSTM-based models exhibit their best performance on our Combined Corpus where both Bi-LSTM and C-LSTM achieve 0.95 accuracy and 0.95 F1-score. CNN and all hierarchical attention models including Conv-HAN maintain a decent performance on this dataset with more than 0.90 accuracy and F1-score. This result indicates that, although neural network-based models may suffer from overfitting for a small dataset (LIAR), they show high accuracy and F1-score on a moderately large dataset (Combined Corpus).

We find that the traditional machine learning models are generally outperformed
by the deep learning models in fake news detection, i.e., the overall accuracy
of the traditional models is much lower than the deep learning ones (Table
\ref{result_table_1}, \ref{result_table_2}). The difference is more prominent on
large dataset, i.e., Combined Corpus which highlights the fact that deep
learning models are prone to overfitting on small dataset. However, despite
being a traditional model, Naive Bayes (with n-gram) shows great promise in fake
news detection which almost reaches the performance of deep learning models and
achieves 93\% accuracy on Combined Corpus. However, further analysis indicates
that the performance of Naive Bayes reaches saturation at some point (2.5K
training data) and after that improves very slowly with the increase of sample
size, while the performance of the deep learning model, i.e., Bi-LSTM has a
greater rate of improvement with the increase of training data (see Figure      
             \ref{rq4_comparison_of_best_models_on_small_part_fake_or_real}). So
it can be deduced that with enough training samples, deep learning models might
be able to outperform Naive Bayes.

\addtocounter{o}{1}
\begin{tcolorbox}[flushleft upper,boxrule=1pt,arc=0pt,left=0pt,right=0pt,top=0pt,bottom=0pt,colback=white,after=\ignorespacesafterend\par\noindent]
\textit{\textbf{Summary of RQ1. How  accurate  are  the  traditional  vs  deep  learning  models  to  detect  fake news?}} 
The deep learning models generally outperform the traditional learning models. \revision{The difference of performance between deep learning and traditional models depends on the dataset length. While deep learning models are vulnerable to overfitting on a small dataset, traditional models like Naive Bayes can show impressive performance on this type of dataset. As the dataset length increases, the performance of the deep learning models also improves, and as a result, the deep learning models outperform the traditional models on a large dataset.}

\end{tcolorbox}


\begin{table*}[ht]
\caption{Performance of Advanced Pre-trained Language Models}
\begin{center}
\resizebox{\columnwidth}{!}
{
\begin{tabular}{lrrrr|rrrr|rrrr}
 & \multicolumn{12}{c}{\textbf{Datasets}} \\ \cmidrule{2-13} 
 & \multicolumn{4}{c}{\textit{\textbf{Liar}}} & \multicolumn{4}{c}{\textit{\textbf{Fake or Real News}}} & \multicolumn{4}{c}{\textit{\textbf{Combined Corpus}}} \\ 
 \midrule
{\textbf{Model}} & \bf{A} & \bf{P} & \bf{R} & \bf{F1} & \bf{A} & \bf{P} & \bf{R} & \bf{F1} & \bf{A} & \bf{P} & \bf{R} & \bf{F1} \\ \midrule
{BERT} & .62 & .62 & .62 & .62 & .96 & .96 & .96 & .96 & .95 & .95 & .95 & .95 \\ \midrule 
{RoBERTa} & \textbf{.62} & \textbf{.63} & \textbf{.62} & \textbf{.62} & \textbf{.98} & \textbf{.98} & \textbf{.98} & \textbf{.98} & \textbf{.96} & \textbf{.96} & \textbf{.96} & \textbf{.96} \\\midrule 
{DistilBERT} & .60 & .60 & .60 & .60 & .95 & .95 & .95 & .95 & .93 & .93 & .93 & .93 \\ \midrule
{ELECTRA} & .61 & .61 & .61 & .61 & .96 & .96 & .96 & .95 & .95 & .95 & .95 & .95 \\ \midrule
{ELMo} & .61 & .61 & .61 & .61 & .93 & .93 & .93 & .93 & .91 & .91 & .91 & .91 \\ 
\bottomrule
\end{tabular}
}
\label{result_table_3}
\end{center}
\end{table*}

\subsection{Can the advanced pre-trained language models outperform the traditional and deep learning models? (RQ2)}
Table \ref{result_table_3} shows the performances of different pre-trained language models on three datasets. While these models incorporate more complex architectures, they do not suffer from overfitting on a smaller dataset as much as the deep learning models do as previously discussed. This is because these models use pre-trained weights in all the layers except the final classification layers. As a result, they do not need a large dataset for fine-tuning their complex architecture. Therefore, all the pre-trained models we evaluated outperform the other traditional ML and deep learning-based models having F1-score no less than 0.62 on the Liar dataset and no less than 0.95 on the Fake or Real News dataset. Given the large dataset (i.e., Combined Corpus), these pre-trained models achieve better performance in the fake news detection task. \revision{We observe that among the pre-trained language models, the BERT and transformer-based models (i.e., BERT, RoBERTa, DistilBERT, ELECTRA) are generally better than the other one (i.e., ELMo). For example, DistillBERT (66M parameters), BERT (110M parameters), Electra (110M parameters), RoBERTa (125M parameters), achieve 0.93, 0.95, 0.95, and 0.96 accuracy, respectively on the Combined Corpus dataset while ELMo (93.6M parameters) achieves 0.91. We also notice that the performance of the transformer-based models is proportionate to their number of pre-trained parameters. This relative performance can be justified by their state-of-the-art results on the text classification task \cite{roberta, distilbert}.}

\addtocounter{o}{1}
\begin{tcolorbox}[flushleft upper,boxrule=1pt,arc=0pt,left=0pt,right=0pt,top=0pt,bottom=0pt,colback=white,after=\ignorespacesafterend\par\noindent]
\textit{\bf{Summary of RQ2. Can the advanced pre-trained language models outperform the traditional and deep learning models?}} In our experiment, the pre-trained models perform significantly better than the traditional and deep learning models on all datasets (Table \ref{result_table_1}, \ref{result_table_2}, \ref{result_table_3}). Since these models are pre-trained to learn contextual text representations on much larger quantities of text corpus and they have produced new state-of-the-art in several text classification tasks \cite{bert_text_classification}, their commanding performance over the traditional and deep learning models in the fake news detection task is quite expected.
\end{tcolorbox}

\subsection{Which model performs best with small training data? (RQ3)}
\begin{figure}[!ht]
\centerline{\includegraphics[width=10cm,height=10cm,keepaspectratio]{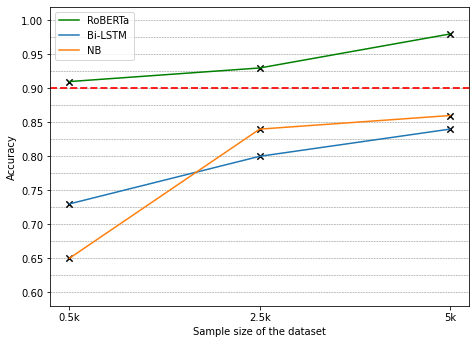}}
\caption{Comparison of Naive Bayes, Bi-LSTM, and RoBERTa with different training dataset size (from Fake or Real News dataset).}
\label{rq4_comparison_of_best_models_on_small_part_fake_or_real}
\end{figure}

We find that pre-trained BERT-based models can perform very well with small datasets. We can realize that from their superior performance on small datasets like Liar and Fake or Real News which is significantly better than other models. To further verify this, we take the best model from each of three types, i.e., Naive Bayes with n-gram (traditional), Bi-LSTM (deep learning), RoBERTa (BERT-based) and compare their performances. As their performances differ on the Fake or Real News dataset by very clear margins, we choose this dataset for this analysis. We report their accuracy on small sets of training data (i.e., 500, 2500, and 5000) chosen from Fake or Real News dataset. We show that RoBERTa achieves notably better performance than the other two (Figure \ref{rq4_comparison_of_best_models_on_small_part_fake_or_real}). RoBERTa reaches more than 90\% accuracy with just 500 training data and continues to improve with the increase of sample size. It hits 98\% accuracy with 5000 sample size. On the other hand, both Naive Bayes and Bi-LSTM perform poorly when the size of training dataset is very small, i.e., 500 (Figure \ref{rq4_comparison_of_best_models_on_small_part_fake_or_real}). Though their performances improve with the increase of dataset size, they fail to achieve 90\% accuracy when the sample size is below 5000.

\begin{figure}[!ht]
\centerline{\includegraphics[width=10cm,height=10cm,keepaspectratio]{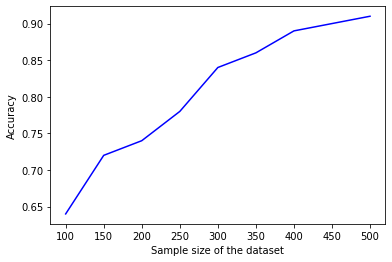}}
\caption{Comparison of RoBERTa's performance on different training dataset size (from Fake or Real News dataset).}
\label{rq4_comparison_roberta_on_small_part_fake_or_real}
\end{figure}

\revision{
We further analyze the performance of RoBERTa on smaller datasets (Figure \ref{rq4_comparison_roberta_on_small_part_fake_or_real}). We find that the model continues to exhibit impressive accuracy (84\%) even when the dataset size is 300. This is because pre-trained weights of RoBERTa have already learned the semantic representation from large text corpora. Fine-tuning on the labeled news articles help to learn the model to distinguish between the real and fake news. We observe that the performance of the model starts to drop quickly after the dataset length has been reduced to less than 300. Reducing the data makes it more difficult for the model to differentiate the news articles. Therefore, the performance decreases quickly.
}

\addtocounter{o}{1}
\begin{tcolorbox}[flushleft upper,boxrule=1pt,arc=0pt,left=0pt,right=0pt,top=0pt,bottom=0pt,colback=white,after=\ignorespacesafterend\par\noindent]
\textit{\textbf{Summary of RQ3. Which model performs best with small training data?}} Pre-trained models (i.e., RoBERTa) show quality performance even with very small training data in our experiment. We find that RoBERTa achieves over 90\% accuracy with a training set of 500 samples only (see Figure \ref{rq4_comparison_of_best_models_on_small_part_fake_or_real}). 
\end{tcolorbox}


\section{Discussion}
\label{sec:discussion}
In this section, we compare the performance of the 19 models we studied along several dimensions like features used, resource requirements, etc. (see Section \ref{sec:discussion-performance}). We then analyze our models' misclassification, which is discussed in Section \ref{sec:discussion-misclassification}. 


\begin{sidewaystable}
\caption{Summary of all models and performances}
\centering

{
\revision{
\begin{tabular}{llllrrr} 
\hline
\multirow{2}{*}{\textbf{Model Type} }                                                             & \multirow{2}{*}{ \textbf{ Model } } & \multirow{2}{*}{\textbf{ Rationale for Picking } }                                                                                                                                                                                       & \multirow{2}{*}{\textbf{ Feature Used } } & \multicolumn{3}{l}{\textbf{ Summary of Result (Acc.) } }                                                                                                                                                                                                \\ 
\cline{5-7}
                                                                                                  &                                     &                                                                                                                                                                                                                                          &                                           & \multicolumn{1}{l}{\textbf{Liar}\textasciitilde{} } & \multicolumn{1}{l}{\begin{tabular}[c]{@{}l@{}}\textbf{Fake}\\\textbf{Or Real} \end{tabular}} & \multicolumn{1}{l}{\begin{tabular}[c]{@{}l@{}}\textbf{ Combined}\\\textbf{Corpus } \end{tabular}}  \\ 
\hline
\multirow{8}{*}{\begin{tabular}[c]{@{}l@{}}Traditional\\Machine\\Learning\\Models \end{tabular}}  & SVM                                 & \multirow{8}{*}{\begin{tabular}[c]{@{}l@{}} These traditional models are\\used in different classification\\tasks including text\\classification. Different\\existing studies used them for\\fake news detection as well. \end{tabular}} & Lexical                                   & 0.56                                                & 0.67                                                                                         & 0.71                                                                                               \\ 
\cline{2-2}\cline{4-7}
                                                                                                  & SVM                                 &                                                                                                                                                                                                                                          & Lexical + Sentiment                       & 0.56                                                & 0.66                                                                                         & 0.71                                                                                               \\ 
\cline{2-2}\cline{4-7}
                                                                                                  & LR                                  &                                                                                                                                                                                                                                          & Lexical + Sentiment                       & 0.56                                                & 0. 67                                                                                        & 0. 76                                                                                              \\ 
\cline{2-2}\cline{4-7}
                                                                                                  & Decision Tree                       &                                                                                                                                                                                                                                          & Lexical + Sentiment                       & 0.51                                                & 0. 65                                                                                        & 0. 67                                                                                              \\ 
\cline{2-2}\cline{4-7}
                                                                                                  & AdaBoost                            &                                                                                                                                                                                                                                          & Lexical + Sentiment                       & 0.56                                                & 0. 72                                                                                        & 0. 74                                                                                              \\ 
\cline{2-2}\cline{4-7}
                                                                                                  & Naïve Bayes                         &                                                                                                                                                                                                                                          & Unigram                                   & 0.60                                                & 0. 82                                                                                        & 0. 91                                                                                              \\ 
\cline{2-2}\cline{4-7}
                                                                                                  & Naïve Bayes                         &                                                                                                                                                                                                                                          & Bigram                                    & 0.60                                                & 0. 86                                                                                        & 0. 93                                                                                              \\ 
\cline{2-2}\cline{4-7}
                                                                                                  & \textit{k}-NN                       &                                                                                                                                                                                                                                          & Empath                                    & 0.54                                                & 0. 71                                                                                        & 0. 71                                                                                              \\ 
\hline
\multirow{6}{*}{\begin{tabular}[c]{@{}l@{}}Deep\\Learning\\Models \end{tabular}}                  & CNN                                 & \begin{tabular}[c]{@{}l@{}}CNN extracts features\\and classify texts by\\transforming words into\\vectors.\end{tabular}                                                                                                                  & \multirow{6}{*}{GloVe embedding}          & 0.58                                                & 0. 86                                                                                        & 0. 93                                                                                              \\ 
\cline{2-3}\cline{5-7}
                                                                                                  & LSTM                                & \begin{tabular}[c]{@{}l@{}} LSTM remembers information\\for long sentences. \end{tabular}                                                                                                                                                &                                           & 0.54                                                & 0. 76                                                                                        & 0. 93                                                                                              \\ 
\cline{2-3}\cline{5-7}
                                                                                                  & Bi-LSTM                             & \begin{tabular}[c]{@{}l@{}} Bi-LSTM analyzes a certain\\part from both\\previous and\\next events. \end{tabular}                                                                                                                         &                                           & 0.58                                                & 0. 85                                                                                        & 0. 95                                                                                              \\ 
\cline{2-3}\cline{5-7}
                                                                                                  & C-LSTM                              & \begin{tabular}[c]{@{}l@{}} Convolutional layer with max-\\pooling combines the local\\features into a global vector to\\help LSTM remembering\\important information. \end{tabular}                                                     &                                           & 0.54                                                & 0. 86                                                                                        & 0. 95                                                                                              \\ 
\cline{2-3}\cline{5-7}
                                                                                                  & HAN                                 & \begin{tabular}[c]{@{}l@{}} HAN applies attention\\mechanism for both\\word-level and sentence-level\\representation. \end{tabular}                                                                                                      &                                           & 0.75                                                & 0. 87                                                                                        & 0. 92                                                                                              \\ 
\cline{2-3}\cline{5-7}
                                                                                                  & Conv-HAN                            & \begin{tabular}[c]{@{}l@{}} Convolutional layer encodes\\embedding into feature for\\word-level and senetence-\\level attention. \end{tabular}                                                                                           &                                           & 0.59                                                & 0. 86                                                                                        & 0. 92                                                                                              \\ 
\hline
\multirow{5}{*}{\begin{tabular}[c]{@{}l@{}}Advanced\\Pre-trained\\Language\\Models \end{tabular}} & BERT                                & \multirow{5}{*}{\begin{tabular}[c]{@{}l@{}}These language models are\textasciitilde{}\\pre-trained on large text corpus\textasciitilde{}\\and can be fine-tuned for\textasciitilde{}\\text classification. \end{tabular}}                 & BERT embeddings                           & 0.62                                                & 0. 96                                                                                        & 0. 95                                                                                              \\ 
\cline{2-2}\cline{4-7}
                                                                                                  & RoBERTa                             &                                                                                                                                                                                                                                          & RoBERTa embeddings                        & 0.62                                                & 0. 98                                                                                        & 0. 96                                                                                              \\ 
\cline{2-2}\cline{4-7}
                                                                                                  & DistilBERT                          &                                                                                                                                                                                                                                          & DistilBERT embeddings                     & 0.60                                                & 0. 95                                                                                        & 0. 93                                                                                              \\ 
\cline{2-2}\cline{4-7}
                                                                                                  & ELECTRA                             &                                                                                                                                                                                                                                          & ELECTRA embeddings                        & 0.61                                                & 0. 96                                                                                        & 0. 95                                                                                              \\ 
\cline{2-2}\cline{4-7}
                                                                                                  & ELMo                                &                                                                                                                                                                                                                                          & ELMo embeddings                           & 0.61                                                & 0. 93                                                                                        & 0. 91                                                                                              \\
\hline
\end{tabular}}
}
\label{table:model_summary}
\end{sidewaystable}

\revision{
\subsection{Analysis of Performance of Different Models}\label{sec:discussion-performance}
In Table \ref{table:model_summary} we summarize the models we studied in our study based on their accuracy across the three datasets, i.e., Liar, Fake or Real, Combined Corpus. Among the eight types of models we 
studied under traditional learning approach, Na\"{i}ve Bayes shows the best accuracy on all the three datasets: combined corpus (0.93), Fake or Real (0.86), and 
Liar (0.60). Among the six traditional deep learning 
models we studied,  there are three different winners in the three datasets: 
C-LSTM shows the best performance on the combined corpus (Accuracy = 0.95), HAN shows the best performance (Accuracy = 0.87) on Fake or Real and HAN shows 
the best performance on the Liar dataset (Accuracy = 0.75). Among the five pre-trained advanced natural 
language deep learning models we studied, RoBERTa shows the best performance across the three datasets: combined corpus (Accuracy = 0.96), 
Fake or Real dataset (Accuracy = 0.98), and Liar (0.62). \bf{Overall, RoBERTa is the best performing model for two datasets (Combined corpus and Fake or Real) 
across all the models we studied, while HAN is the best performer for the Liar dataset.} 
}   




\revision{
\bf{The performance of Na\"{i}ve Bayes (with n-gram) is only slightly less than the deep learning and pre-trained language models.} As such, Na\"{i}ve Bayes can be a good choice for fake news detection on a
sufficiently large dataset with hardware constraint. Naive Bayes (with n-gram) has also been reported to show good
performance in spam detection in earlier studies \cite{new3}. We find that the
performance of Naive  Bayes (with n-gram) is almost equivalent to the
performances of deep learning models on Combined Corpus (see Table
\ref{result_table_1}). Hence, in the absence of hardware resource requirement of
deep learning and advanced pre-trained models (a possible case for non-profit
blogs/websites), Naive Bayes with n-gram can be a suitable option with
a sufficiently large dataset. Note that the required size of the dataset may vary
with its nature, i.e., the number of topics included. However, Naive Bayes fails to
achieve considerable accuracy when trained on a minimal sample set (see
Figure \ref{rq4_comparison_of_best_models_on_small_part_fake_or_real}).}
\revision{
Among the diverse features, we studied for the traditional learning models (lexical, sentiment, n-grams), bigram-based 
models (e.g., Na\"{i}ve Bayes) show better performance than other features. Overall, the incorporation of sentiment indicators into the 
models did not improve their performance. For example, for SVM the performance is the same (0.71) for both settings: lexical and lexical + sentiment. 
Therefore, \textbf{Sentiment features are not observed as useful for fake news
detection in our study.} The
classification of news (as real or fake) has very little to do with the
polarity (i.e., sentiment), as fake news can be made up in both directions
(positive or negative).
}
\revision{
\bf{While two LSTM-based models (Bi-LSTM, C-LSTM) are the best performer among all the traditional deep learning models, their 
performance degrades significantly when the dataset sizes are smaller (see RQ3)}.
We observe that LSTM based models show gradual improvement when the dataset
length increases from LIAR to Combined Corpus. The more an article contains
information, the less these models will be vulnerable to overfitting, and the
better they will perform. Hence, neural network-based models may show high
performance on a larger dataset over 100k samples \cite{new2}. 
}
\revision{
\bf{The pre-trained BERT-based models outperform the other models not only on the overall datasets but also on smaller samples of the datasets (see RQ3)}. We see that the
BERT-based model (i.e., RoBERTa) is capable of achieving high accuracy (over
90\%) even with a limited sample size of 500 data (see Figure
\ref{rq4_comparison_of_best_models_on_small_part_fake_or_real}). Hence, these
models can be utilized for fake news detection in different languages where
a large collection of labeled data is not feasible. Different pre-trained BERT
models are already available for different languages, e.g., ALBERTO for Italian
\cite{italian_bert}, AraBERT for Arabic \cite{arabic_bert}, BanglaBERT
\footnote{\url{https://github.com/sagorbrur/bangla-bert}}. 
}

\revision{
We measured the average training time (per epoch) and GPU usage
(during testing) for each BERT-based model on Combined Corpus. We find that the
training time needed by DistilBERT is almost half of BERT and RoBERTa, and it
requires less GPU for testing (i.e., prediction) as well (see Table
\ref{gpu_bert_models}). Therefore, \bf{while DistilBERT shows 0.93 accuracy on the combined corpus which is only slightly behind BERT (0.95) or RoBERTa (0.96), DistilBERT can be useful for production-level usage with
hardware constraint and less response time.} 
This is because DistilBERT is developed using the concept of knowledge
distillation \cite{distil_knowledge, distil_model}. Hence, it is suitable for production-level usage considering its' high
performance and low resource requirement. 
}

\begin{table}[t]
\caption{Comparison of training time and GPU usage (in testing) for BERT-based models}
\resizebox{\columnwidth}{!}{%
\begin{tabular}{lrp{4cm}p{4cm}}
\toprule
\textbf{Model} & \textbf{\#Parameters} & \textbf{Avg. training time per epoch (sec)} & \textbf{GPU used in testing (GB)} \\ \midrule
DistilBERT & 66 M & 2175 & 2.48 \\ 
BERT & 110 M & 3149 & 2.95 \\ 
RoBERTa & 125 M & 4020 & 3.07 \\ 
\bottomrule
\end{tabular}%
}
\label{gpu_bert_models}
\end{table}

%

\subsection{Misclassification Analysis}\label{sec:discussion-misclassification}
\begin{figure}[t]
\centerline{\includegraphics[height=9cm,keepaspectratio]{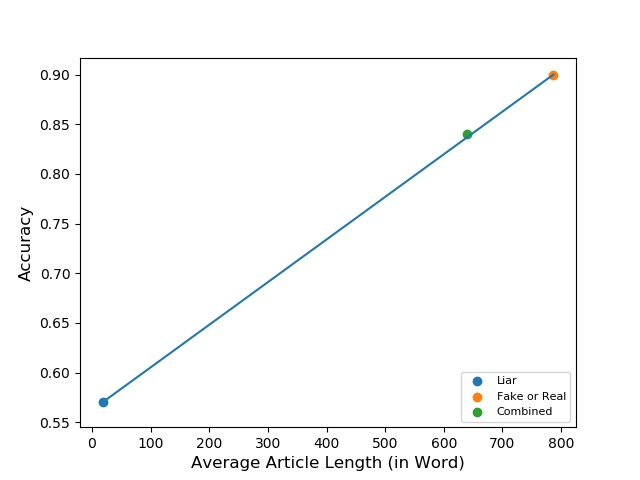}}
\caption{Relation between models' performance and article length.}
\label{article_len_fig}
\end{figure}
\revision{
Among the three datasets in our study, the best models (pre-trained language models) show more than 96\% accuracy for two datasets (Combined corpus and 
Fake or Real). For the other dataset (Liar), the best performing model was  HAN with 75\% accuracy. Compared to the other two datasets, 
the Liar dataset has significantly smaller articles (18 words on average) compared to the other two datasets (average 644 words for Combined Corpus and 765 words for Fake or Real news). 
Indeed, we have observed that when the number
of training data is constant, the accuracy of this model is proportional to the
average article length of news (see Figure~\ref{article_len_fig}). We confirmed this 
by analyzing the performance of the Na{i}ve Bayes model on 5000
randomly selected records from each of our three datasets. This observation is also consistent 
with other models. Thus, with the
increase of news article length, the models can become more
accurate, because those can extract more information to classify the
news correctly.
}

Among the three datasets, two datasets are related to politics (Fake or Real news, Liar), while the other dataset (Combined Corpus) has fake news about diverse topics like health and research, politics, economy, and so on (see \fig\ref{dataset_fig_1_LDA_Topic} in 
Section \ref{sec:materials_and_methods}). To understand whether the topic of the news has any effect on the classification, we apply topic-based analysis on the
fake news articles from the Combined corpus, which our model misclassifies as real. 
We then map each misclassified case to the ten topics that we found in \fig\ref{dataset_fig_1_LDA_Topic} of the combined corpus.  
Overall, quotes are greatly misused to design fake news. We find that
the most frequent words in these articles are `said', `study', and `research'.
The profuse use of the word `said' indicates how fake news sources misconstrue
quotes to make these as believable as possible and carry out their own
agendas.

\begin{table}[!ht]
\caption{Topic-wise percentage of false positive news in the Combined Corpus}
\begin{center}

\scalebox{1.4}
{
\begin{tabular}{lr}
\toprule
\textbf{Topic}                                                 & \textbf{False Positive News (\%)} \\ \midrule
Health and Research & 49.6                                                                        \\ 
Politics                                                       & 27.6                                                                        \\ 
Miscellaneous                                                  & 22.8                                                                        \\ \bottomrule
\end{tabular}%
}
\label{discussion_table_3}
\end{center}
\end{table}

\revision{
The topic-wise analysis of misclassification in the combined corpus shows that 49.6\% of the false positive news
(that are mispredicted as fake in our study) are related to health and
research-based topics (Table \ref{discussion_table_3}). On the other hand, a
tiny portion (27.6\%) of the false positive news are related to politics. This
high false positive rate of health and research-related news bears evidence that
clickbait news on health and research can be produced more convincingly. A
slight change in the actual research article will still keep the fake news in
the close proximity of the actual article, which makes it difficult to identify
them as fake news. In this way, it is quite easy for clickbait news sources to
attract people by publishing news claiming the invention of a vaccine for
incurable diseases like terminal cancer. Hence, although in recent times the
media has focused mostly on combating unauthentic political news, it should also
pay attention to stop the proliferation of false health and research-related
news for public safety. We can realize this lesson even better if we think of
the impact of fake news during the current COVID-19 pandemic. Corona related
fake news has caused serious troubles and confusion among the people. Several
fake news such as ``Alcohol cures COVID-19",``5G spreads coronavirus", etc have
affected people both physically and
mentally\footnote{\url{https://www.bbc.com/news/stories-52731624}, Accessed on:
Oct 05, 2020.}. Considering the threats associated with it, corona related fake
news has been compared to a second pandemic or
infodemic\footnote{\url{https://www.nature.com/articles/d41586-020-01409-2},
Accessed on: Oct 05, 2020.}. 
}






\section{Conclusions}
\label{sec:conclusions}
In this study, we present an overall performance analysis of 19 different machine
learning approaches on three different datasets. Eight out of the 19 models are traditional learning models, six models are traditional deep learning models, and five models are advanced pre-trained language models like BERT. 
We find that BERT-based models have achieved better performance
than all other models on all datasets. More importantly,
we find that pre-trained BERT-based models are robust to the size of the dataset and can
perform significantly better on very small sample size. 
We also find that Naive Bayes with n-gram can attain similar results to
neural network-based models on a dataset when the dataset size is sufficient.
The performance of LSTM-based models greatly depends on the length of the
dataset as well as the information given in a news article. With adequate
information provided in a news article, LSTM-based models have a higher
probability of overcoming overfitting. 
The results and findings based on our comparative analysis can
facilitate future researches in this direction and also help the organizations
(e.g., online news portals and social media) to choose the most suitable model
who are interested in detecting fake news. Our future work in this direction will focus 
on designing models that can detect misinformation and health-related fake news that are prevalent in social media during the COVID-19 pandemic.
%




\bibliographystyle{plain}
\bibliography{References}


\end{document}